# A COMPARATIVE ANALYSIS OF SEMICONDUCTOR WAFER MAP DEFECT DETECTION WITH IMAGE TRANSFORMER

Sushmita Nath

*Chittagong University of Engineering and Technology*

*Abstract*— Predictive maintenance is an important sector in modern industries which improves fault detection and cost reduction processes. By using machine learning algorithms in the whole process, the defects detection process can be implemented smoothly. Semiconductor is a sensitive maintenance field that requires predictability in work. While convolutional neural networks (CNNs) such as VGG-19, Xception and Squeeze-Net have demonstrated solid performance in image classification for semiconductor wafer industry, their effectiveness often declines in scenarios with limited and imbalanced data. This study investigates the use of the Data-Efficient Image Transformer (DeiT) for classifying wafer map defects under data-constrained conditions. Experimental results reveal that the DeiT model achieves highest classification accuracy of 90.83%, outperforming CNN models such as VGG-19(65%), SqueezeNet(82%), Xception(66%) and Hybrid(67%). DeiT also demonstrated superior F1-score (90.78%) and faster training convergence, with enhanced robustness in detecting minority defect classes. These findings highlight the potential of transformer-based models like DeiT in semiconductor wafer defect detection and support predictive maintenance strategies within semiconductor fabrication processes.

*Keywords*— Convolutional Neural Network, Data Efficient Image Transformer, Semiconductor wafer maps, Deep learning, Comparative analysis

## I. INTRODUCTION

Predictive maintenance (PdM) technique uses data analysis, sensors and machine learning (ML) to predict errors in equipment [3]. It also uses artificial intelligence (AI), an enormous sector of ML, to make the process easy for predicting faults in devices [2]. It helps to identify faults and errors early. It can prevent any major and minor losses of the industries, which will ultimately save costs for the industry. It will increase the equipment's lifetime by indicating its failure beforehand. It is also saving the workers time and life by preventing accidents caused by sudden breakdowns or failures.

This research presents a comprehensive analysis of semiconductor wafer defect detection by using machine learning and transformer-based architecture. The study primarily utilized the WM-811k augmented and preprocessed semiconductor wafer image dataset, which contains wafer map images categorized into nine distinct defect types. To ensure a fair and systematic evaluation, the dataset was manually partitioned into training, validation and testing subsets. Alongside the Data-efficient image Transformer(DeiT), several other models were developed and assessed, including VGGNet, Xception, SqueezeNet and a custom hybrid model [1]. Each model's performance was evaluated using key classification metrics such as accuracy, precision, recall, F1-score, and confusion matrix analysis, following either training from scratch or fine-tuning with appropriate configurations.

### A. Objectives

The basic objectives of the research on predictive maintenance in semiconductor wafer map manufacturing industries are given below:

- To implement and evaluate transformer-based model and machine learning models for semiconductor wafer defect detection,
- To assess model performance using comprehensive classification matrics,
- To demonstrate the advantage of transformer-based model over CNNs and identify the most effective architecture for automated defect classification.

These objectives of the research will help to improve semiconductor wafer manufacturing by reducing equipment failures, improving defect detection and enhancing maintenance efficiency.

### B. Scopes

Transformer-based image defects detection is important for many industries like manufacturing, industry 4.0, healthcare and medical diagnostics, smart cities and transportation, retail and customer experience, finance and banking, agriculture and environment monitoring. In the manufacturing sector, it is used for early fault detection in motors and bearings. Many applications of this model exist in semiconductor manufacturing. These applications are defect detection, quality control, predictive maintenance, equipment failure prevention, yield optimization, process control, supply chain optimization, logistics, AI-driven semiconductor design and electronic design automation. Many companies like TSMC, ASML, Intel, Samsung, NVIDIA, AMD, Google's TPU design uses AI-driven model to operate effortlessly.

## II. LITERATURE REVIEW

Previous researches related to this research activity are briefly discussed in this section.

### A. Previous Studies

Bhatnagar et al. [4] decided to introduce and contrast different transfer learning techniques that may identify the right flaws in the wafer map. Using the WM811k real-world wafer map dataset, they examine the performance of several transfer learning-based models, including VGG19, MobileNet, ResNet, and DenseNet. They have grouped the dataset's eight wafer map defect classes into four groups.

Their tests have shown that VGG19 has the best accuracy on the test data, with a 95.56% accuracy rate.

To enhance the learning of discriminative features for complex wafer defects, this study by Batool et al. [5], propose an Attention-Augmented Convolutional Neural Network (A2CNN) model. A2CNN incorporates both channel and spatial attention mechanisms, along with a global average pooling layer to reduce overfitting and a focused loss function to minimize misclassification. The model is evaluated on the MixedWM38 wafer defect dataset using 10-fold cross-validation, demonstrating high generalization capability. It achieves outstanding performance with accuracy, precision, recall, and F1-score of 98.66%, 99.0%, 98.55%, and 98.82%, respectively, surpassing previous approaches in detecting intricate mixed-type defects.

In research by Iandola et al. [6], deep neural networks has largely focused on improving accuracy, yet many architectures now offer comparable performance. Smaller models, however, provide distinct advantages such as reduced communication overhead during distributed training and lower bandwidth usage when deploying updates to edge devices like autonomous systems. These compact architectures are also well-suited for implementation on memory-constrained hardware platforms like FPGAs. In this context, SqueezeNet was developed to deliver AlexNet-level accuracy on ImageNet using 50× fewer parameters and can be compressed to under 0.5 MB, making it over 500× smaller than AlexNet.

### B. Research Gap

Most existing studies primarily use traditional CNN models like VGGNet and ResNet for classifying semiconductor defects, often overlooking newer, more data-efficient approaches. Although Vision Transformers (ViT) have achieved strong performance in general computer vision tasks, they are still underutilized in wafer defect detection, especially in data scenarios like those in WM811k augmented and preprocessed dataset. There is also a lack of research that systematically compares CNNs, transformers, and hybrid methods on the same dataset. Few studies have conducted controlled evaluations to assess the relative strengths of these different architectures. Moreover, transformer models designed for efficiency, such as DeiT, are rarely tested against larger CNNs in wafer classification tasks. This highlights a gap in validating whether these efficient models can outperform more complex alternatives in this specific domain. Deit uses the self-attention mechanism, which makes it better suited for real-world wafer datasets.

## III. METHODOLOGY

This research was conducted through the implementation of various methods, organizing multiple approaches to ensure a comprehensive analysis. Various techniques and strategies were used to gather and interpret data. This chapter covers various aspects, including data collection, approaches for selecting relevant features, strategies for training predictive models, and techniques for evaluating model performance. Fig.1. illustrates the comprehensive methodology employed for wafer defect classification, including data collecting, preprocessing, model training, and evaluation.

### A. Data Collection

To effectively predict defects using ML, various types of data will be required. For this study, the widely used WM811k augmented and preprocessed wafer defect dataset was selected, which is publicly available for research in semiconductor manufacturing [7]. This dataset contains images representing various defect categories such as Center, Donut, Edge-Local, Edge-Ring, Local, Near-Full, Random, Scratch, and None. It comprises 9000 wafer map images, each labeled based on the type of defect present. These labeled images make it highly suitable for supervised learning applications, particularly for defect detection and classification. Equal number of samples from each defect category results in a balanced dataset for model training and evaluation.

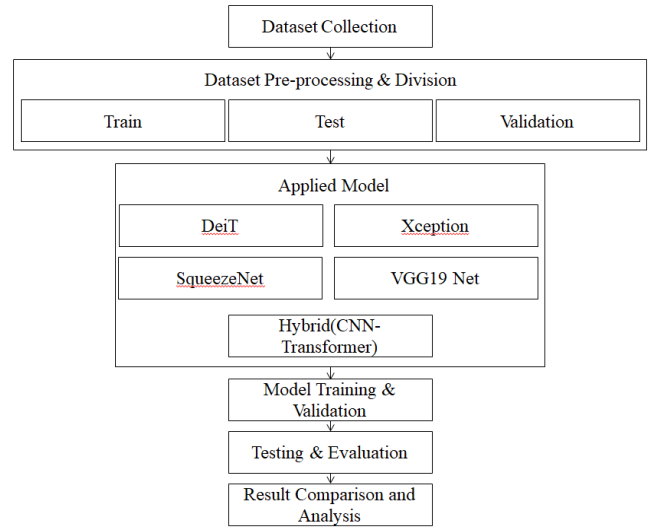

Fig. 1. Methodology Flowchart for the research

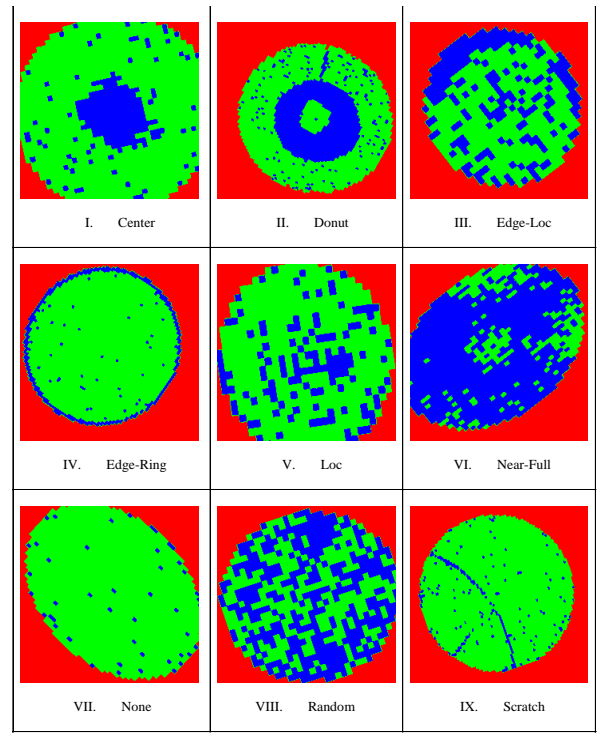

Fig. 2. Examples of different defect types from each class

Each image in Fig. 2 reflects a unique pattern commonly identified during the post-fabrication inspection process. Nine defect categories are:

I. Center: Shows a concentrated cluster of defects near the wafer's center, often associated with central process disruptions.

II. Donut: Reveals a circular band of anomalies, with relatively fewer faults at the center and periphery, typically linked to alignment deviations or uneven etching.

III. Edge-Loc (Edge Localized): Indicates faults limited to one side of the wafer, generally caused by edge related processing problems or handling damage.

IV. Edge-Ring: Displays a ring of defects along the wafer's border, usually due to non-uniform edge deposition or etching inconsistencies.

V. Local: Characterized by defects limited to a small, irregular zone, often resulting from localized contamination or particle interference.

VI. Near-Full: Depicts widespread defects covering a majority of the wafer surface, typically signifying major process malfunctions or equipment failures.

VII. None: Represents wafers with minimal or no defects, serving as the control class during model training and validation.

VIII. Random: Shows scattered defects across the wafer without a recognizable structure, commonly originating from random environmental noise or inspection artifacts.

IX. Scratch: Features straight, scratch-like patterns caused by mechanical abrasion, mishandling, or contact during processing.

*B. Evaluation Metrics*

To estimate the performance of these techniques on predictive maintenance, we will need to utilize some evaluation metrics including accuracy, precision, recall, F1-score, confusion matrix.

Accuracy: Accuracy is calculated by using the following formula:

$$Accuracy = \frac{(TP+TN)}{(TP+FN+FP+TN)} \ldots\ldots\ldots\ldots\ldots\ldots(1)$$

TP, TN, FN, FP are important values in evaluating performance of model. Here, TP indicates true positive, TN indicates true negative, FN indicates false negative and FP indicates false positive. The more the value, the better will be the accuracy. It is basically used in the classification sector. In a multi-class setting, it becomes

$$Accuracy = \frac{\text{Number of correct predictions}}{\text{Total Predictions}} \ldots\ldots\ldots\ldots(2)$$

Accuracy becomes less reliable when the dataset is imbalanced.

Precision: Precision, or Positive Predictive Value, measures how many of the predicted positive cases are actually correct. It becomes especially important when the cost of false positives is high. In essence, it reflects the reliability of a model's positive predictions.

$$Precision = \frac{TP}{(TP+FP)} \ldots\ldots\ldots\ldots\ldots\ldots(3)$$

In multi-class classification, precision is calculated separately for each class. It can be averaged in two common ways: macro averaging, which treats all classes equally, and weighted averaging, which accounts for the number of actual samples in each class. These methods provide an overall view of model performance across all categories.

Recall: Recall, also known as sensitivity or the true positive rate, measures how well a model identifies actual positive instances. It reflects the ability of the classifier to capture most of the true positives. A higher recall score means fewer positive cases are missed by the model.

$$Recall = \frac{TP}{(TP+FN)} \ldots\ldots\ldots\ldots\ldots\ldots\ldots (4)$$

F1-Score: The F1-Score is the harmonic mean of Precision and Recall, providing a single metric that balances both. It is especially useful in scenarios with imbalanced class distributions. A high F1-Score indicates that both Precision and Recall are performing well.

$$F1 - Score = \frac{2*Precision*Recall}{Precision+Recall} \ldots\ldots\ldots\ldots\ldots (5)$$

When minimizing both false positives and false negatives is important, a balanced evaluation is needed. The F1 score is especially useful for imbalanced datasets, where accuracy might seem high but performance on minority classes is weak. In such cases, F1 gives a clearer picture of the model's true effectiveness.

Confusion Matrix: A confusion matrix is a square chart that summarizes the performance of a classification model. Each row corresponds to the actual class, while each column shows the predicted class. It helps identify both correct and incorrect predictions for each category individually. A confusion matrix shows how well a model performs for each individual class. For nine classes, it is displayed as a 9 by 9 grid where the diagonal entries indicate correct predictions for each class. Values outside the diagonal represent misclassifications.

IV. RESULT AND EVALUATION

*A. Model Performance*

This thesis made use of several machine learning models including DeiT, Xception, SqueezeNet, VGG-19 and a hybrid model.

DeiT: The DeiT model achieved the highest accuracy of 90.83%, surpassing all CNN-based models. Its data efficient training and self-attention mechanisms allowed it to generalize well across both common and rare defect classes.

It demonstrated strong performance in terms of precision, recall, and achieved the top F1-score among all models.

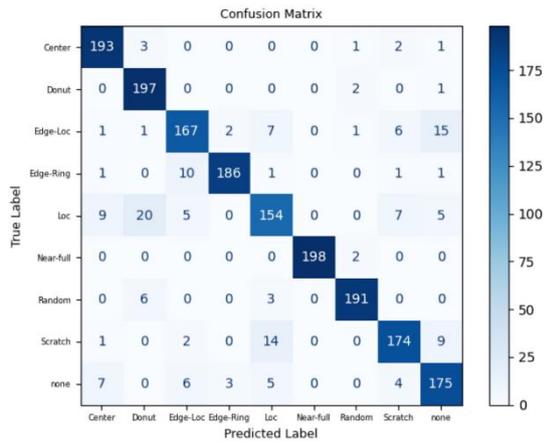

Fig. 3. Confusion Matrix of DeiT

Fig. 4. Bar chart comparing precision, recall, and F1 across all classes of DeiT

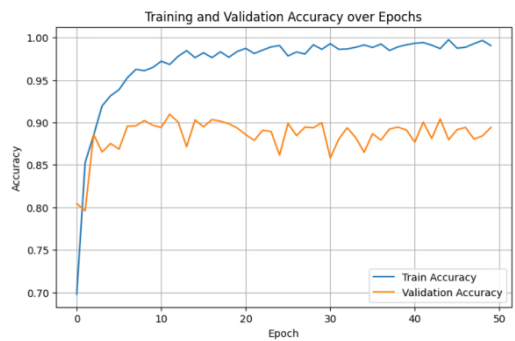

Fig. 5. Training and Validation Accuracy of DeiT

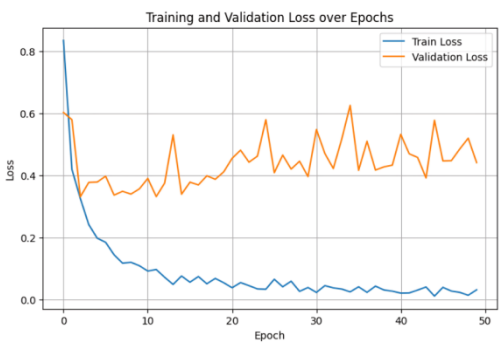

Fig. 6. Training and Validation Loss of DeiT

A detailed evaluation of the performance of all the trained models for classifying wafer defects are presented here.

TABLE I. The Comparison Table of Implemented Models

| | Accuracy | Precision | Recall | F1-Score | Average Time/Step(s) |
|---|---|---|---|---|---|
| DeiT | 0.9083 | 0.9089 | 0.9083 | 0.9078 | 0.035 |
| Xception | 0.66 | 0.6303 | 0.6556 | 0.6361 | 0.090 |
| SqueezeNet | 0.82 | 0.84 | 0.82 | 0.82 | 0.092 |
| VGG-19 | 0.65 | 0.73 | 0.65 | 0.61 | 0.263 |
| Hybrid | 0.67 | 0.68 | 0.67 | 0.67 | 0.20 |

## Conclusion

This study carried out a comprehensive comparison of various deep learning models for classifying defects in semiconductor wafers using the WM-811k augmented and preprocessed dataset. The models explored included well-established Convolutional Neural Networks (such as VGGNet, Xception, and SqueezeNet), a Transformer based model (DeiT), and a combined CNN-Transformer architecture. Performance evaluation was based on key classification metrics—accuracy, precision, recall, F1-score—and analysis of the confusion matrix. Among all the models, DeiT stood out with the highest accuracy (90.83%) and maintained consistent results across all defect types, including rare and underrepresented ones. Although CNN models like Xception and SqueezeNet handled common defect types reasonably well, they struggled with infrequent classes. The hybrid strategy provided moderate improvements in stability but came with added training complexity and only marginal accuracy gains. These findings underscore the potential of vision transformer models as a reliable solution for wafer defect detection. Additionally, the research shed light on how different architectural features influence classification performance. It addressed a crucial gap in the field by applying a data-efficient transformer to the semiconductor defect domain, demonstrating its ability to outperform both traditional CNNs and more complex hybrid models.

The future work on this topic can be – Analyzing real time data for predicting defects more accurately, Focusing on using different model to see which one is better and Developing a fully automated system for the maintenance of manufacturing industries

## Acknowledgment

I would like to express my heartfelt gratitude to my honourable teachers for their constant guidance. The guidance was truly informative and helpful for me. Their effort for teaching and research created tough opportunities to develop my scientific expertise and my increasing interest in the mechatronics and industrial engineering field.